\documentclass[11pt]{article}
\usepackage[final]{acl}
\usepackage{times}
\usepackage{latexsym}
\usepackage[T1]{fontenc}
\usepackage[utf8]{inputenc}
\usepackage{microtype}
\usepackage{inconsolata}

\usepackage{amsmath,amsfonts,bm}

\def\eqref#1{equation~\ref{#1}}
\def\1{\bm{1}}

\DeclareMathAlphabet{\mathsfit}{\encodingdefault}{\sfdefault}{m}{sl}
\SetMathAlphabet{\mathsfit}{bold}{\encodingdefault}{\sfdefault}{bx}{n}

\usepackage{booktabs}
\usepackage{graphicx}
\usepackage{algorithm}
\usepackage{algorithmic}
\usepackage{amssymb}
\usepackage{float}
\usepackage{xcolor}
\usepackage{multirow}
\usepackage[normalem]{ulem}
\usepackage{xspace}
\usepackage{tikz}
\usepackage{booktabs}
\usepackage{multirow}
\usepackage{makecell}
\usetikzlibrary{arrows.meta,backgrounds,calc,matrix,positioning}

\newif\ifshowauthornotes
\showauthornotesfalse

\newcommand{\Hc}{H_c}
\newcommand{\block}{B}
\newcommand{\mask}{\texttt{[MASK]}}
\usepackage{xspace}

\newcommand{\kbcomment}[1]{\ifshowauthornotes{\color{cyan}KB: #1}\fi}

\newcommand{\DStill}{\textsc{DStill}\xspace}
\newcommand{\DFlash}{DFlash }

\title{\DStill: Trajectory Consistency Distillation for Accelerated One-Step Drafting in Block Diffusion Speculative Decoding}

\author{Shreyas Kulkarni$^\dagger$\thanks{Co-first authors. $^\dagger$Corresponding author. Email: \texttt{shreyas.kulkarni2@capitalone.com}.}, Uttam Gurram$^*$, Madhav Kanda, Krishna Teja Chitty Venkata,\\ \textbf{Shichu Chen, Kartik Balasubramaniam, Supriyo Chakraborty, Sambit Sahu}\\
AI Foundations, Capital One}

\begin{document}

\maketitle
% Block-diffusion speculative decoding accelerates large language model inference by drafting multiple tokens in parallel. However, one-step drafting in discrete diffusion is limited by token factorization error: each token is predicted from the same masked block, without conditioning on the tokens selected at other positions in that step. This can produce inconsistent drafts and reduce acceptance length. Iterative refinement mitigates this limitation by conditioning later predictions on previously committed tokens, though additional passes increase drafting cost. To address this trade-off between draft fidelity and inference latency, we introduce \textsc{DStill}, a two-stage framework that distills a target-conditioned drafter's own multi-pass refinement into one-step drafting. We first train the drafter to complete partially revealed blocks\ugcomment{improving its iterative refinement}, then use its refinement trajectories to train a student to predict the final completion in one pass. Resampling intermediate states across epochs encourages the student to predict the same completion from different partial inputs. Fully masked states are also sampled throughout training, providing direct supervision for one-step drafting at inference time. Across reasoning, math, and code benchmarks, \textsc{DStill} improves acceptance length by up to $10\%$ over one-step \DFlash and achieves an \textcolor{red}{XX}$\times$ speedup over autoregressive decoding.\ugcomment{"Resampling ... one-step deployment" points seem a bit too technical for abstract?}
\begin{abstract}
Block-diffusion speculative decoding accelerates large language model inference by drafting multiple tokens in parallel. However, one-step drafting in discrete diffusion is limited by token factorization error: each token is predicted from the same masked block, without conditioning on the tokens selected at other positions in that step. This can produce inconsistent drafts and reduce acceptance length. Iterative refinement mitigates this limitation by conditioning later predictions on previously committed tokens, though additional passes increase drafting cost. To address this trade-off between draft fidelity and inference latency, we introduce \DStill, a two-stage framework that distills a target-conditioned drafter's own multi-step refinement into one-step drafting. We first train the drafter to complete partially revealed blocks, then use its refinement trajectories to train a student to predict the final completion in one pass. Using a curriculum that gradually transitions from partial to fully masked inputs, we successfully transfer the multi-step refinement capabilities into one step. Across reasoning, math, code benchmarks on three model families \DStill improves acceptance length by up to $28\%$ and speedup by up to $25\%$ over one-step \DFlash, achieving up to $6.45\times$ speedup over auto-regressive decoding.
\end{abstract}

\section{Introduction}
\label{sec:intro}

\input{figure1}

Speculative decoding accelerates Large Language Model (LLM) inference by using a small, inexpensive draft model to propose future tokens and verifying those proposals against the target model in a single parallel forward pass \citep{leviathan2023fast,chen2023accelerating}. Under the standard rejection-sampling verifier, this procedure is lossless as it preserves the output distribution of autoregressive sampling from the target model. An effective drafter must achieve a high \emph{acceptance length} $\tau$, the expected number of draft tokens accepted per verification step, while keeping drafting latency low.

Target-conditioned\footnote{A target-conditioned drafter
uses hidden states computed by the target LLM as additional
context for predicting draft tokens, rather than relying
solely on the preceding token sequence. The target model
remains frozen.} block-diffusion drafters, such as
\DFlash \citep{chen2026dflash}, predict an entire block
in parallel. Each block starts in the \emph{root state}: the latest verified token is visible as an anchor, and all future-token positions are masked. A one-pass drafter predicts each masked position without conditioning on the tokens selected at the other masked positions. Independent predictions can combine tokens from incompatible continuations (Figure~\ref{fig:factorization}a) \citep{lee2026flow}. Refining the block over multiple denoising passes progressively commits high-confidence tokens, fixing their values for subsequent passes. The remaining positions can then condition on these commitments and improve their agreement with the target. However, the additional drafter passes increase latency, and the benefit of longer accepted prefixes may not offset this cost.

We introduce \DStill, a two-stage trajectory consistency distillation framework that transfers the benefits of multi-step refinement to one-step speculative drafting. In the first stage, we train a target-conditioned drafter to complete partially masked blocks in a single forward pass. After training, we use this drafter as a teacher to collect multi-step denoising trajectories, recording the intermediate partially unmasked blocks and final completions which are fully unmasked which we refer to as \emph{endpoints}. In the second stage, a student initialized from the teacher learns to predict each trajectory's final completion directly from its recorded states. At inference time, the student drafts a block in one forward pass, followed by standard target-model verification  (Figure~\ref{fig:dstill-overview}).

During distillation, we sample a state from the teacher's recorded trajectory and use it as input to the student. We resample this state across epochs while keeping the trajectory's final completion fixed as the training target. This exposes the student to different partial commitments for the same block, with varying amounts of context for predicting the remaining tokens. To supervise one-step drafting, we use a curriculum that gradually increases \emph{root-bias}, in which we increase the probability of sampling the root state through a schedule as we progress through training. 

Our procedure adapts the theoretical foundation of Consistency Models \cite{song2023consistency} to discrete diffusion by learning to map any intermediate point on a masking trajectory directly to the same endpoint. Just as these continuous models enforce a self-consistency property where outputs remain invariant for arbitrary points along a given probability flow trajectory, we train the student to map partially revealed intermediate blocks to a constant, fully refined target sequence. By internalizing this discrete refinement trajectory into its hidden states, the student learns to sharpen the blurry averaging of independent marginal probabilities (Figure~\ref{fig:factorization}b,c), mirroring how continuous consistency models learn to convert pure noise into coherent data samples using a single step. We make three main contributions:

% On these inputs, the student learns to predict the teacher's refined completion in a single forward pass. Shared endpoint supervision encourages the student to favor the teacher's selected continuation while retaining a factorized output distribution , analogous to sequence-level distillation for non-autoregressive translation \citep{zhou2020understanding}.

\paragraph{Transferring Multi-Step Refinement to One-Step Drafting.} We introduce \DStill, a framework that distills a drafter's own multi-step refinement process into a one-step speculative drafter. By training the student to predict the teacher's refined completion from various states along its trajectory, we successfully transfer multi-step capabilities to a single-step process. This approach recovers $35.8\%$ and $28.6\%$ of the multi-step teacher's acceptance length improvements on GPT-OSS-120B and on Nemotron-3.5-30B-A3B respectively.
\paragraph{Zero-Overhead Efficiency Gains Across Models and Domains.} We evaluate \DStill across three target-model families on benchmarks spanning reasoning, math, code, and chat. Compared to DFlash, our method achieves up to $28\%$ higher acceptance length and up to $25\%$ wall-clock speedup. Crucially, after offline distillation, these gains are realized using the exact same drafter architecture, avoiding test-time latency penalties associated with additional correction heads or tree-search decoding.
\paragraph{Generalization to Block Diffusion Speculators.} We also apply trajectory distillation to DSpark's block diffusion backbone \citep{cheng2026dspark}, showing a $4.5\%$ improvement in acceptance length (Appendix~\ref{appendix:dspark}) demonstrating its general applicability and motivating extensions to other block diffusion speculators.

\section{Related Work}
\label{sec:related}

\paragraph{Speculative Decoding.} Speculative decoding \citep{leviathan2023fast,chen2023accelerating} accelerates autoregressive LLM generation by having a cheap draft model propose several future tokens, which the target model then verifies in a single forward pass; a rejection-sampling scheme guarantees that the resulting output distribution exactly matches that of the target model. The approach builds on earlier parallel-decoding ideas such as blockwise parallel decoding \citep{stern2018blockwise}. Follow-up work extends verification from a single chain to a candidate tree \citep{miao2023specinfer,chen2024sequoia}, while draft-free variants such as Jacobi-style lookahead decoding \citep{fu2024lookahead} remove the draft model entirely at the cost of a weaker acceptance guarantee.

\paragraph{Feature-based and Multi-head Drafting.} EAGLE \citep{li2024eagle} and its successors \citep{li2024eagle2,li2025eagle3} draft tokens autoregressively using target-model features and embeddings of preceding tokens. Medusa \citep{cai2024medusa} instead attaches multiple prediction heads to the target backbone, predicting future tokens independently in parallel. Hydra \citep{hydra2024} extends this design by conditioning each draft head on preceding draft tokens, introducing sequential dependencies within the proposed continuation while retaining lightweight prediction heads.

\paragraph{Parallel and Diffusion Drafters.} \DFlash \citep{chen2026dflash} replaces sequential autoregressive drafting with a lightweight block-diffusion model that predicts an entire block in one pass while conditioning on hidden features from the target model. DFlare \citep{zhang2026dflare} scales this design by equipping each draft layer with a learned mixture over a broader set of target-layer features, increasing draft capacity. Other work restores intra-block dependencies through lightweight auxiliary modules: Domino \citep{huang2026domino} applies a causal head to the preliminary distributions produced by a parallel backbone, and DSpark \citep{cheng2026dspark} couples a parallel backbone with a small sequential markov head while adapting the verification length based on confidence estimates. DDTree \citep{ringel2026ddtree} keeps predictions parallel but uses their per-position distributions to construct a draft tree for joint verification by the target model.

% \textsc{DStill}\ugcomment{, on the other hand, follows a two-stage training in which it trains teacher with partially masked blocks to improve its iterative refinement, followed by training (just mentioning it here for a better flow)} trains a one-pass backbone against the endpoints of an iterative denoising
% teacher. The procedure requires a parallel diffusion drafter that can score partially masked blocks and expose its denoising trajectory. This permits different backbone designs in principle; empirical claims are limited to the configurations evaluated here.
% We also apply trajectory distillation to DSpark's block-diffusion backbone and report the results in Appendix Table~\ref{tab:dsparkplusdstill}. We leave the evaluation of trajectory distillation with DFlare, Domino, and DDTree to future work.

\paragraph{Masked and discrete diffusion language models.} Absorbing-state (masked) diffusion
formulates generation as iterative unmasking with weights derived from an evidence lower bound (ELBO)
\citep{austin2021d3pm,lou2024sedd,sahoo2024mdlm},
and large masked-diffusion language models such as LLaDA \cite{llada2025} and Dream \citep{dream2025} trade sampling steps for
generation quality via confidence-based remasking.  

% Block Diffusion combines autoregressive
% generation across blocks with diffusion within each block, reusing a key--value (KV) cache across
% already-generated blocks \citep{arriola2025bd3lm}. Our block structure, an anchored,
% fixed-size block that attends bidirectionally within itself and causally to everything before it,
% is a direct instance of this design, specialized to the drafting setting. Fast-dLLM also
% identifies token dependence as a limitation of parallel unmasking and uses confidence-aware
% decoding to address it \citep{wu2025fastdllm}. 

\paragraph{Consistency and few-step distillation.} Consistency models learn a map from any point
on a diffusion trajectory to its endpoint, collapsing many-step sampling into one or a few forward
passes \citep{song2023consistency}, later generalized to arbitrary source-target
pairs along the trajectory \citep{kim2024ctm}. In the discrete/masked setting,
self-distillation through time (SDTT) transfers multi-step unmasking distributions to fewer
steps \citep{deschenaux2025sdtt}; \citet{zhang2026t3d} distill a many-step
masked-diffusion model through generative-trajectory matching, and \citet{kim2026cdlm} combine
consistency-guided distillation with cache-friendly block causality to accelerate masked-diffusion
inference. Consistency LLMs fine-tune autoregressive models to predict fixed points from
states along Jacobi-decoding trajectories, with an auxiliary ground-truth loss
\citep{kou2024cllm}. These methods accelerate generation by the language model being
distilled. 

% \textsc{DStill} operates within speculative decoding: a lightweight, target-conditioned
% drafter learns from its own refinement trajectories while the autoregressive target remains
% fixed. Like SDTT, we supervise each position with the teacher distribution from the pass that
% committed its token (\emph{committing-pass supervision}). Our training procedure combines
% partial-mask teacher training with reuse of recorded trajectories and root-biased state sampling to
% transfer refinement gains into one-pass speculative proposals. Intermediate inputs are
% resampled across epochs, while the anchor-only root state receives direct supervision
% throughout. The objective matches discrete endpoint tokens and distributions.

\section{\DStill}
\label{sec:method}

Figure~\ref{fig:dstill-overview} describes \DStill.
In Stage I, we train a target-conditioned drafter to complete
partially revealed \emph{reference blocks} from sequences generated
by the autoregressive target model. The trained drafter then serves
as the teacher. We freeze its parameters and record its multi-step
refinement trajectories offline. In Stage II, a student initialized
from the teacher learns to predict the teacher's final completions
from states along these trajectories. At inference, the student
drafts a block in a single forward pass, and the target LLM verifies the proposed tokens.

\subsection{Setup and notation}
\label{sec:prelim}

\begin{figure*}[t]
\centering
\begin{tikzpicture}[
  x=1cm,y=1cm,
  font=\sffamily\fontsize{9}{10.5}\selectfont,
  heading/.style={
    font=\sffamily\bfseries\fontsize{9}{10.5}\selectfont,
    anchor=west
  },
  note/.style={
    font=\sffamily\fontsize{8}{9.5}\selectfont,
    align=center
  },
  token/.style={
    draw=black!25,
    rounded corners=0.6mm,
    minimum height=0.42cm,
    inner sep=0pt,
    font=\sffamily\fontsize{9}{10.5}\selectfont
  },
  masked/.style={fill=black!4,text=black!65},
  visible/.style={
    fill=teal!12,
    draw=teal!50!black,
    text=teal!55!black
  },
  flow/.style={
    -{Latex[length=1.5mm,width=1.1mm]},
    line width=0.65pt,
    draw=black!55
  }
]
\path[use as bounding box] (0,0) rectangle (15.9,5.38);

% Shared context.
\node[
  draw=black!18,
  fill=blue!3,
  rounded corners=0.8mm,
  text width=15.34cm,
  minimum height=0.48cm,
  inner xsep=2mm,
  inner ysep=1mm,
  align=center,
  font=\sffamily\fontsize{9}{10.5}\selectfont
] at (7.95,5.08)
  {\textbf{Context:} \emph{Visiting the Golden Gate from \ldots}};

\draw[black!15] (5.20,0.12) -- (5.20,4.66);
\draw[black!15] (11.80,0.12) -- (11.80,4.66);

% (a) Joint probabilities versus the product of token marginals.
\node[heading] at (0.05,4.51) {(a) Token factorization};
\node[note,text=black!65] at (2.575,4.11)
  {Two valid city pairs (toy)};

\node[
  anchor=west,
  font=\sffamily\bfseries\fontsize{8}{9.5}\selectfont
] at (0.14,3.66) {Pair};
\node[
  font=\sffamily\bfseries\fontsize{8}{9.5}\selectfont
] at (3.60,3.66) {Joint};
\node[
  font=\sffamily\bfseries\fontsize{8}{9.5}\selectfont
] at (4.60,3.66) {Product};

\draw[black!35,line width=0.45pt] (0.08,3.44) -- (5.07,3.44);

\foreach \yy in {3.14,2.66}
  \fill[teal!9] (0.08,\yy-0.20) rectangle (5.07,\yy+0.20);
\foreach \yy in {2.18,1.70}
  \fill[orange!14] (0.08,\yy-0.20) rectangle (5.07,\yy+0.20);

\node[anchor=west] at (0.14,3.14) {San Francisco};
\node[anchor=west] at (0.14,2.66) {Mountain View};
\node[anchor=west,text=orange!65!black] at (0.14,2.18)
  {San View};
\node[anchor=west,text=orange!65!black] at (0.14,1.70)
  {Mountain Francisco};

\foreach \yy/\pp/\qq in {
  3.14/0.60/0.36,
  2.66/0.40/0.16,
  2.18/0/0.24,
  1.70/0/0.24
} {
  \node at (3.60,\yy) {$\pp$};
  \node at (4.60,\yy) {$\qq$};
}

\node[note,text=orange!65!black] at (2.575,1.06)
  {Incoherent samples: $48\%$};
\node[note,text=black!65] at (2.575,0.51)
  {Exact marginals can still\\produce incompatible pairs.};

% (b) Different recorded states from the same teacher trajectory.
% The trajectory commits San, then punctuation, then Francisco.
\begin{scope}[xshift=0.75cm]
\node[heading] at (4.70,4.51)
  {(b) Shared endpoint across epochs};
\node[note,text=black!65] at (7.82,4.11)
  {One stored teacher trajectory};

\node[note] at (5.15,3.42) {Epoch 1};
\node[note,text=blue!65!black] at (5.15,2.60)
  {Epoch 2\\(root)};
\node[note] at (5.15,1.78) {Epoch 3};

\node[token,visible,minimum width=0.70cm] at (6.18,3.42)
  {San};
\node[token,masked,minimum width=1.35cm] at (7.25,3.42)
  {[M]};
\node[token,visible,minimum width=0.55cm] at (8.28,3.42)
  {.};

\node[token,masked,draw=blue!55,minimum width=0.70cm]
  at (6.18,2.60) {[M]};
\node[token,masked,draw=blue!55,minimum width=1.35cm]
  at (7.25,2.60) {[M]};
\node[token,masked,draw=blue!55,minimum width=0.55cm]
  at (8.28,2.60) {[M]};

\node[token,visible,minimum width=0.70cm] at (6.18,1.78)
  {San};
\node[token,masked,minimum width=1.35cm] at (7.25,1.78)
  {[M]};
\node[token,masked,minimum width=0.55cm] at (8.28,1.78)
  {[M]};

\node[
  draw=teal!60!black,
  fill=teal!7,
  rounded corners=1mm,
  minimum width=1.65cm,
  minimum height=2.06cm,
  align=center,
  font=\sffamily\bfseries\fontsize{9}{11}\selectfont
] (student) at (9.86,2.60) {Student\\$D_\phi$};

\draw[flow] (8.61,3.42) -- (9.01,3.21);
\draw[flow] (8.61,2.60) -- (student.west);
\draw[flow] (8.61,1.78) -- (9.01,1.99);

\node[note,text=teal!55!black] at (7.16,0.98)
  {Fixed teacher endpoint};
\node[token,visible,minimum width=0.70cm] at (6.18,0.47)
  {San};
\node[token,visible,minimum width=1.35cm] at (7.25,0.47)
  {Francisco};
\node[token,visible,minimum width=0.55cm] at (8.28,0.47)
  {.};

\draw[flow,draw=teal!65!black]
  (student.south) -- (9.86,0.47) -- (8.65,0.47);
\node[note,text=teal!55!black,anchor=west] at (9.97,1.01)
  {one\\pass};
\end{scope}

% (c) Illustrative concentration toward the selected endpoint.
% Before: (0.6,0.4) at each position; after: (0.9,0.1).
% Incompatible mass changes from 2*0.6*0.4 to 2*0.9*0.1.
\node[heading] at (12.00,4.51)
  {(c) Illustrative sharpening};
\node[note,text=black!65] at (13.95,4.11)
  {Root probabilities (illustrative)};

\draw[black!45,line width=0.45pt]
  (12.20,1.70) -- (15.72,1.70);
\draw[black!45,line width=0.45pt]
  (12.20,1.70) -- (12.20,3.54);

\node[note,anchor=east] at (12.11,1.70) {$0$};
\node[note,anchor=east] at (12.11,3.45) {$1$};

\foreach \xx in {12.90,14.85} {
  \fill[black!30]
    (\xx-0.35,1.70) rectangle (\xx-0.02,2.75);
  \fill[teal!65]
    (\xx+0.08,1.70) rectangle (\xx+0.41,3.275);
  \node[note,text=black!65] at (\xx-0.185,2.94) {$0.6$};
  \node[note,text=teal!60!black] at (\xx+0.245,3.47) {$0.9$};
}

\node[note] at (12.92,1.42) {San};
\node[note] at (14.87,1.42) {Francisco};

\fill[black!30] (12.48,1.04) rectangle (12.66,1.22);
\node[note,anchor=west] at (12.73,1.13) {Before};
\fill[teal!65] (14.10,1.04) rectangle (14.28,1.22);
\node[note,anchor=west] at (14.35,1.13) {After};

\node[note,text=orange!65!black] at (13.95,0.51)
  {Incoherent samples\\$48\% \;\longrightarrow\; 18\%$};
\end{tikzpicture}

\caption{\textbf{Factorization and distillation.}
(a) \emph{Joint} shows a toy autoregressive target's token-pair probabilities.
\emph{Product} multiplies its per-position marginals, corresponding to
independent sampling and assigning 48\% probability to incoherent pairs.
(b) Recorded states resampled across epochs share one endpoint.
(c) Toy probabilities illustrate concentration toward that endpoint;
the student remains factorized.
[M] denotes masking.
The anchor is omitted in this figure.}
\label{fig:factorization}
\end{figure*}
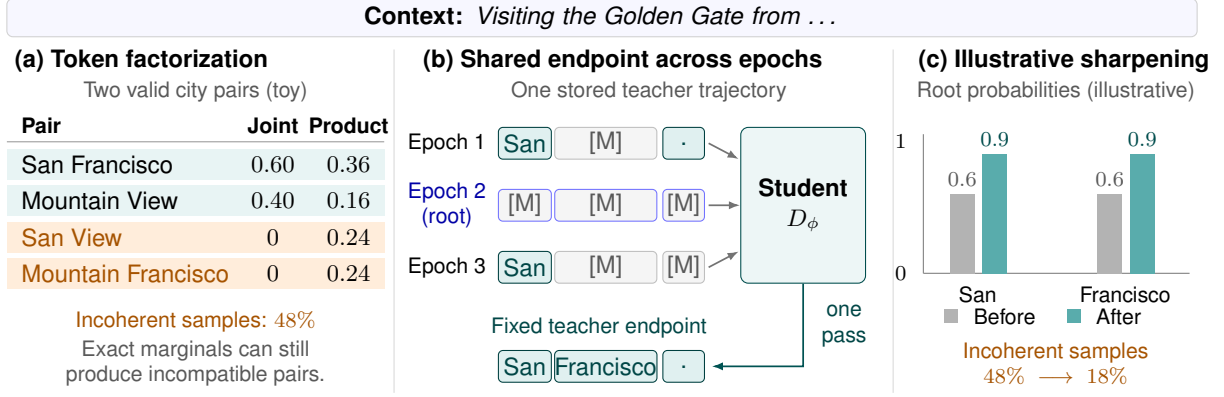

Let $a$ be the latest verified token, and let $\Hc$ be the target
model's cached hidden states for the tokens before $a$.
The teacher and student operate on blocks of $\block$ positions,
with the fixed anchor $a$ at position $0$ and future tokens at
positions $1,\ldots,\block-1$.
In the \emph{root state}, $x_1=(a,\mask,\ldots,\mask)$,
all future-token positions are masked. Following DFlash \citep{chen2026dflash}, the drafter $D_\theta$
attends bidirectionally within each block $x$ and to target
context $\Hc$ strictly before the anchor. It predicts a vocabulary
distribution $p_{\theta,i}(\cdot\mid x,\Hc)$ at each masked
position $i$. Independently sampling the future tokens from
the root predictions gives the proposal distribution
\begin{equation}
  \label{eq:one-step-proposal}
  q_\theta(y_{1:\block-1}\mid a,\Hc)
  = \prod_{i=1}^{\block-1}
    p_{\theta,i}(y_i\mid x_1,\Hc).
\end{equation}
Greedy drafting instead takes each position's top-1 prediction.
Figure~\ref{fig:factorization}a illustrates how factorized
sampling can combine incompatible continuations.

Multi-step refinement allows later predictions to condition on
tokens selected earlier. The drafter completes a block over
$K$ passes, selecting the highest-confidence masked positions
and committing their predicted tokens at each pass.
We measure confidence by the largest token probability at
each position. Once committed, a token remains fixed for
the remaining passes.

We denote the inputs to these passes by $x_1,\ldots,x_K$,
where $x_1$ is the root state. The completed block after the
final pass is denoted by $x_0$ and called the \emph{endpoint}.
Appendix~\ref{sec:prelim-details} gives the reveal schedule
and verification procedure.

% Vector schematic. All probabilities are illustrative, not measured results.
% The epoch inputs are nested states from one teacher trajectory, resampled
% in a nonchronological order; they are not independently remasked endpoints.

\subsection{Stage I: Partial-mask teacher training}
\label{sec:method-partial-mask}

In Stage I, we train the drafter to complete partially revealed
blocks in a single forward pass. This prepares the drafter for
multi-step refinement, where later passes use previously
committed tokens to predict the remaining tokens.

For a reference block $z=(a,z_1,\ldots,z_{\block-1})$, we sample
a reveal probability $r$ uniformly from a discrete set.
We choose these probabilities so that training blocks contain,
on average, as many visible tokens as the teacher sees at
each refinement pass.

Given $r$, we reveal each future-token position independently
with probability $r$. Revealed positions contain their reference
tokens, and the remaining positions contain $\mask$, producing
the input block $x^r$. The expected fraction of masked future
positions, or \emph{corruption level}, is therefore $1-r$.
The anchor remains visible, and $r=0$ gives the root state.
We resample the reveal probability and positions across epochs.

Given this partially revealed block, we train the drafter to
recover the reference tokens at the remaining masked positions.
Let $M$ denote these positions. For $|M|>0$, the unweighted
training objective is
\begin{equation}
  \label{eq:teacher-main}
  \mathcal L_{\mathrm{teacher}}
  = -\frac{1}{|M|}
    \sum_{i\in M}
    \log p_{\theta,i}(z_i\mid x^r,\Hc).
\end{equation}
The reveal probabilities and optional loss weights are specified
in Appendix~\ref{sec:teacher-details}. After this training,
we use the drafter's own predictions to collect the refinement
trajectories needed for student distillation.

\subsection{Trajectory collection}
\label{sec:method-collect}

We freeze the trained drafter as the teacher and run its
confidence-guided refinement offline, once per training block.
Each rollout starts from the root state and progressively
reveals the teacher's predictions. These predictions may differ
from the reference tokens used in Stage I.
We use greedy token selection for all models: at each selected
position, the teacher commits its highest-probability token
(Appendix~\ref{sec:gpu_compute}).

We record each trajectory so that its states and predictions
can be reused during student training. Each record contains
the final block $x_0$ and the pass at which each token was
committed. At each pass, we also store the teacher's top-$k$
token IDs and probabilities at the masked positions.

Because committed tokens are never revised, the endpoint and
commit passes are sufficient to reconstruct every input state.
To reconstruct a state, we reveal endpoint tokens committed
before that pass and mask the remaining future positions.
The anchor remains visible throughout.

The reconstructed states serve as inputs to the student.
We supervise the remaining masked positions using the teacher's
committed tokens and the probability distributions recorded
when those tokens were committed. Record formats, reconstruction,
and the collection algorithm are presented in
Appendix~\ref{sec:trajectory-details}.

\subsection{Stage II: One-step trajectory consistency distillation}
\label{sec:method-distill}

The student $D_\phi$ shares the teacher's architecture and is
initialized from its weights. During distillation, we sample
a state from the teacher's recorded trajectory and use it
as input to the student. The student is trained to predict
the trajectory's final completion at the positions that
remain masked.

Different input states from the same trajectory share the same
final completion. For each position that remains masked,
the target token and teacher distribution stay fixed across
these states. We refer to this shared supervision as
\emph{trajectory consistency}.
Figure~\ref{fig:factorization}b,c illustrates how this supervision
can increase the probabilities of tokens from the teacher's
selected completion. Incompatible combinations can then become
less likely while the student's output distribution remains
factorized.

\paragraph{Root-biased state sampling.}
To expose the student to different amounts of visible context,
we train on both root and intermediate states.
Intermediate states provide committed tokens that help predict
the remaining positions, while root states match the input
used for one-step drafting. With probability $p_{\mathrm{root}}$,
we sample the root state $x_1$; otherwise, we uniformly sample
a recorded intermediate state. Blocks without intermediate
states use root.

As training progresses, we increase $p_{\mathrm{root}}$ from
$0.5$ to $0.85$ using a cosine schedule, establishing a curriculum that places greater
emphasis on the root state later in training (Appendix~\ref{sec:root-details}).
The input state is resampled across epochs, while trajectory's
final completion and teacher distributions remain fixed.

\paragraph{Endpoint supervision.}
After selecting an input state, we supervise the student's
predictions at the positions that remain masked. The training
loss mask restricts supervision to response-token positions,
and the anchor is excluded.

Each supervised position receives two training signals:
the token committed by the teacher and the teacher's probability
distribution from the pass that committed it. Cross-entropy
(CE) trains the student to predict the committed token,
while knowledge distillation (KD) transfers the teacher's
uncertainty.

To compute the KD term, we use the teacher's retained top-$k$
tokens and group the remaining probability mass into a single
category. Let $Q_{\mathrm{end},i}$ denote this distribution
for position $i$. Let $P_{\phi,i}$ denote the student's
distribution given $(x_t,\Hc)$, using the same teacher-selected
token IDs and the same aggregate category.

For each supervised position $i$, we combine CE and KD with
weight $\alpha\in[0,1]$ on the KD term:
\begin{equation}
  \label{eq:distill-loss}
  \begin{aligned}
    \mathcal L_i
      &= (1-\alpha)
      \underbrace{
        [-\log p_{\phi,i}(x_{0,i}\mid x_t,\Hc)]
      }_{\text{endpoint CE}} \\
      &\quad+\alpha
      \underbrace{
        D_{\mathrm{KL}}
        (Q_{\mathrm{end},i}\,\|\,P_{\phi,i})
      }_{\text{top-}k\text{ KD}}.
  \end{aligned}
\end{equation}
We use $\alpha=0.2$ and $k=16$, and average over the positions
included in the loss.

We optimize this objective while keeping the teacher fixed.
Both drafters reuse the target's token embeddings and
language-model (LM) head, which projects hidden states to
vocabulary logits. These shared components also remain frozen,
so Stage II updates only the remaining student parameters.
Appendix~\ref{sec:loss-details} gives the exact probability
construction, optional trajectory weights, batch aggregation,
and training algorithm.

\subsection{Inference}
\label{sec:method-deploy}

\begin{table*}[t]
\centering
\caption{Generation speedup $S$ and mean acceptance length $\tau$ across target models and benchmarks at concurrency $C=1$, temperature = $0$, reasoning = on/medium. Params denotes drafter parameters in Millions.} 
% Reduce space between columns to allow for larger text size
\setlength{\tabcolsep}{2.5pt} 
\resizebox{\textwidth}{!}{
\begin{tabular}{ll c cccccccccccccc}
\toprule
\multirow{2}{*}{Model} & \multirow{2}{*}{Method} & \multirow{2}{*}{Params (M)} & \multicolumn{2}{c}{GSM8K} & \multicolumn{2}{c}{MATH-500} & \multicolumn{2}{c}{HumanEval} & \multicolumn{2}{c}{MBPP} & \multicolumn{2}{c}{MT-Bench} & \multicolumn{2}{c}{SWE-Bench} & \multicolumn{2}{c}{\textsc{Average}} \\
\cmidrule(lr){4-5} \cmidrule(lr){6-7} \cmidrule(lr){8-9} \cmidrule(lr){10-11} \cmidrule(lr){12-13} \cmidrule(lr){14-15} \cmidrule(lr){16-17}
& & & $S$ & $\tau$ & $S$ & $\tau$ & $S$ & $\tau$ & $S$ & $\tau$ & $S$ & $\tau$ & $S$ & $\tau$ & $S$ & $\tau$ \\
\midrule
\multirow{2}{*}{GPT-OSS-120B} 
& DFlash & $785$ & 4.81$\times$ & 6.3 & 4.01$\times$ & 5.17 & 3.53$\times$ & 4.58 & 3.57$\times$ & 4.63 & 2.42$\times$ & 3.09 & 2.33$\times$ & 2.96 & 3.45$\times$ & 4.45 \\
& DStill (Ours) & $785$ & \textbf{5.17}$\times$ & \textbf{6.79} & \textbf{4.27}$\times$ & \textbf{5.54} & \textbf{3.7}$\times$ & \textbf{4.83} & \textbf{3.8}$\times$ & \textbf{4.97} & \textbf{2.6}$\times$ & \textbf{3.34} & \textbf{2.53}$\times$ & \textbf{3.25} & \textbf{3.68}$\times$ & \textbf{4.79} \\
\midrule
\multirow{2}{*}{Gemma-4-31B}
& DFlash  & $1536$ & 5.85$\times$ & 6.78 & 5.9$\times$ & 6.84 & 5.2$\times$ & 6.03 & 4.3$\times$ & 5.00 & 3.16$\times$ & 3.67 & 2.74$\times$ & 3.17 & 4.53$\times$ & 5.25 \\
& DStill (Ours) & $1536$ & \textbf{6.45}$\times$ & \textbf{7.48} & \textbf{6.28}$\times$ & \textbf{7.28} & \textbf{5.59}$\times$ & \textbf{6.49} & \textbf{4.76}$\times$ & \textbf{5.52} & \textbf{3.47}$\times$ & \textbf{4.03} & \textbf{2.92}$\times$ & \textbf{3.38} & \textbf{4.91}$\times$ & \textbf{5.70} \\
\midrule
\multirow{2}{*}{Gemma-4-26B-A4B}
& DFlash  & $430$ & 2.86$\times$ & 4.57 & 2.98$\times$ & 4.65 & 2.58$\times$ & 4.09 & 2.23$\times$ & 3.55 & 1.78$\times$ & 2.81 & 1.62$\times$ & 2.54 & 2.34$\times$ & 3.70 \\
& DStill (Ours) & $430$ & \textbf{3.42}$\times$ & \textbf{5.57} & \textbf{3.53}$\times$ & \textbf{5.59} & \textbf{3.11}$\times$ & \textbf{5.02} & \textbf{2.79}$\times$ & \textbf{4.53} & \textbf{2.08}$\times$ & \textbf{3.38} & \textbf{1.75}$\times$ & \textbf{2.81} & \textbf{2.78}$\times$ & \textbf{4.48} \\
\midrule
\multirow{2}{*}{Nemotron-3.5-30B-A3B}
& DFlash  & $481$ & 2.62$\times$ & 4.81 & 2.66$\times$ & 4.86 & 2.31$\times$ & 4.27 & 2.08$\times$ & 3.79 & 1.77$\times$ & 3.18 & 1.62$\times$ & 2.89 & 2.18$\times$ & 3.97 \\
& DStill (Ours) & $481$ & \textbf{2.98}$\times$ & \textbf{5.62} & \textbf{3.02}$\times$ & \textbf{5.64} & \textbf{2.57}$\times$ & \textbf{4.88} & \textbf{2.33}$\times$ & \textbf{4.36} & \textbf{1.94}$\times$ & \textbf{3.6} & \textbf{1.66}$\times$ & \textbf{3.04} & \textbf{2.42}$\times$ & \textbf{4.52} \\
\bottomrule
\end{tabular}
}
\label{tab:model_performance}
\end{table*}

% \textbf{NOTE: }Gemma-4's global-attention layers use head\_dim 512, which is served only by the FlashAttention-4 path intended for Blackwell; on Hopper (H100) this runs with elevated per-decode-step latency and no faster full-CUDA-graph alternative, inflating the speculative verify cost and raising Gemma-4's break-even acceptance length relative to models with standard head dimensions (e.g., GPT-OSS-120B, Nemotron-3.5-30B-A3B). This is a serving-kernel limitation of running Gemma-4 on Hopper, not a limitation of the DFlash/DSpark/DStill method

At inference, the student receives the target's cached hidden
states $\Hc$ and the root state $x_1$. A single forward pass
produces the proposed block
$\hat x_0=(a,\hat y_1,\ldots,\hat y_{\block-1})$.

For independently sampled one-pass proposals, the student's
per-position distributions provide the proposal conditionals
required by the standard rejection-sampling verifier.
This verifier preserves the target model's sampling distribution.
Greedy verification accepts tokens that match the target's
top-1 predictions.

After a rejection, the target supplies a correction token.
If all draft tokens are accepted, it supplies a bonus token.
This target-generated token becomes the next anchor,
starting another drafting and verification step.
At fixed one-pass drafting cost, longer accepted prefixes can
improve end-to-end speedup; Section~\ref{sec:experiments}
evaluates resulting wall-clock trade-off.

\section{Experiments}
\label{sec:experiments}

Table~\ref{tab:model_performance} presents the main comparison across target models and
benchmarks, reporting generation speedup and acceptance length.
We also measure how much of the teacher's eight-pass refinement gain transfers to one-pass
drafting and examine the teacher objective, student supervision, and state-sampling curriculum.

\begin{table}[t]
\centering
\caption{Generation speedup $S$ (relative to standard autoregressive decoding) for GPT-OSS-120B across benchmarks and request concurrencies $C \in \{1,8,16,32\}$, at temperature $0$ and $1$, reasoning\_effort $=$ medium. Avg~$\tau$ is the mean acceptance length averaged over the four concurrencies. For each dataset, temperature, and concurrency, faster method is in \textbf{bold}.}
\setlength{\tabcolsep}{3pt}
\renewcommand{\arraystretch}{1.1}
\resizebox{\columnwidth}{!}{
\begin{tabular}{ll ccccc ccccc}
\toprule
\multirow{3}{*}{Dataset} & \multirow{3}{*}{Method} & \multicolumn{5}{c}{\textit{Temperature $=0$}} & \multicolumn{5}{c}{\textit{Temperature $=1$}} \\
\cmidrule(lr){3-7} \cmidrule(lr){8-12}
 & & \multirow{2}{*}{Avg $\tau$} & \multicolumn{4}{c}{Speedup $S$ ($\times$)} & \multirow{2}{*}{Avg $\tau$} & \multicolumn{4}{c}{Speedup $S$ ($\times$)} \\
\cmidrule(lr){4-7} \cmidrule(lr){9-12}
 & & & $C{=}1$ & $C{=}8$ & $C{=}16$ & $C{=}32$ & & $C{=}1$ & $C{=}8$ & $C{=}16$ & $C{=}32$ \\
\midrule
\multirow{2}{*}{GSM8K} & DFlash & 6.3 & 4.81 & 3.01 & 3.04 & 3.59 & 5.54 & 4.17 & 2.05 & 2.55 & 3.3 \\
 & DStill & 6.86 & \textbf{5.17} & \textbf{3.26} & \textbf{3.29} & \textbf{3.92} & 5.93 & \textbf{4.54} & \textbf{2.21} & \textbf{2.75} & \textbf{3.42} \\
\cmidrule(lr){1-12}
\multirow{2}{*}{MATH-500} & DFlash & 5.16 & 4.01 & 2.47 & 2.53 & 3 & 4.37 & 3.29 & 2.15 & 2.09 & 2.49 \\
 & DStill & 5.51 & \textbf{4.27} & \textbf{2.67} & \textbf{2.62} & \textbf{3.06} & 4.78 & \textbf{3.69} & \textbf{2.34} & \textbf{2.16} & \textbf{2.71} \\
\cmidrule(lr){1-12}
\multirow{2}{*}{HumanEval} & DFlash & 4.54 & 3.53 & 2.15 & 2.19 & 2.61 & 4.11 & 3.12 & 1.93 & 1.9 & 2.39 \\
 & DStill & 4.88 & \textbf{3.7} & \textbf{2.31} & \textbf{2.38} & \textbf{2.69} & 4.4 & \textbf{3.33} & \textbf{2.07} & \textbf{2.05} & \textbf{2.46} \\
\cmidrule(lr){1-12}
\multirow{2}{*}{MBPP} & DFlash & 4.7 & 3.57 & 2.26 & 2.29 & 2.65 & 4.25 & 3.25 & 2 & 1.97 & 2.4 \\
 & DStill & 4.98 & \textbf{3.8} & \textbf{2.39} & \textbf{2.42} & \textbf{2.69} & 4.58 & \textbf{3.48} & \textbf{2.15} & \textbf{2.12} & \textbf{2.56} \\
\cmidrule(lr){1-12}
\multirow{2}{*}{MT-Bench} & DFlash & 3.1 & 2.42 & 1.53 & 1.52 & 1.86 & 2.92 & 2.25 & 1.44 & 1.37 & 1.76 \\
 & DStill & 3.31 & \textbf{2.6} & \textbf{1.64} & \textbf{1.63} & \textbf{1.98} & 3.05 & \textbf{2.39} & \textbf{1.47} & \textbf{1.42} & \textbf{1.77} \\
\cmidrule(lr){1-12}
\multirow{2}{*}{SWE-Bench} & DFlash & 2.9 & 2.33 & 1.43 & 1.41 & 1.62 & 2.66 & 2.06 & 1.29 & 1.24 & 1.5 \\
 & DStill & 3.19 & \textbf{2.53} & \textbf{1.52} & \textbf{1.55} & \textbf{1.77} & 2.84 & \textbf{2.2} & \textbf{1.37} & \textbf{1.33} & \textbf{1.55} \\
\bottomrule
\end{tabular}
}

\label{tab:gptoss_concurrency}
\end{table}

\begin{table}[t]
\centering
\caption{Generation speedup $S$ (relative to standard autoregressive decoding) for the Nemotron-3.5-30B-A3B trained drafters across benchmarks and request concurrencies $C \in \{1,8,16,32\}$, for reasoning on/off and temperature $0$/$1$.}
\setlength{\tabcolsep}{3pt}
\renewcommand{\arraystretch}{1.1}
\resizebox{\columnwidth}{!}{
\begin{tabular}{ll ccccc ccccc}
\toprule
\multirow{3}{*}{Dataset} & \multirow{3}{*}{Method} & \multicolumn{5}{c}{\textit{Temperature $=0$}} & \multicolumn{5}{c}{\textit{Temperature $=1$}} \\
\cmidrule(lr){3-7} \cmidrule(lr){8-12}
 & & \multirow{2}{*}{Avg $\tau$} & \multicolumn{4}{c}{Speedup $S$ ($\times$)} & \multirow{2}{*}{Avg $\tau$} & \multicolumn{4}{c}{Speedup $S$ ($\times$)} \\
\cmidrule(lr){4-7} \cmidrule(lr){9-12}
 & & & $C{=}1$ & $C{=}8$ & $C{=}16$ & $C{=}32$ & & $C{=}1$ & $C{=}8$ & $C{=}16$ & $C{=}32$ \\
\midrule
\multicolumn{12}{c}{\textit{Reasoning $=$ ON}} \\
\midrule
\multirow{2}{*}{GSM8K} & DFlash & 4.81 & 2.62 & 1.42 & 2.48 & 1.98 & 4.18 & 2.42 & 1.16 & 2.04 & 1.7 \\
 & DStill & 5.71 & \textbf{2.98} & \textbf{1.67} & \textbf{2.79} & \textbf{2.19} & 4.73 & \textbf{2.64} & \textbf{1.3} & \textbf{2.34} & \textbf{1.9} \\
\cmidrule(lr){1-12}
\multirow{2}{*}{MATH-500} & DFlash & 4.77 & 2.66 & 1.36 & 2.35 & 2.06 & 4.06 & 2.3 & 1.16 & 1.97 & 1.78 \\
 & DStill & 5.52 & \textbf{3.02} & \textbf{1.52} & \textbf{2.69} & \textbf{2.39} & 4.53 & \textbf{2.47} & \textbf{1.28} & \textbf{2.19} & \textbf{1.95} \\
\cmidrule(lr){1-12}
\multirow{2}{*}{HumanEval} & DFlash & 4.29 & 2.31 & 1.19 & 2.05 & 1.89 & 4.21 & 2.35 & 1.17 & 1.99 & 1.81 \\
 & DStill & 4.89 & \textbf{2.57} & \textbf{1.33} & \textbf{2.33} & \textbf{2.13} & 4.74 & \textbf{2.63} & \textbf{1.27} & \textbf{2.24} & \textbf{2} \\
\cmidrule(lr){1-12}
\multirow{2}{*}{MBPP} & DFlash & 3.78 & 2.08 & 1.01 & 1.77 & 1.64 & 3.76 & 2.11 & 1.02 & 1.7 & 1.57 \\
 & DStill & 4.37 & \textbf{2.33} & \textbf{1.14} & \textbf{2.03} & \textbf{1.9} & 4.29 & \textbf{2.33} & \textbf{1.15} & \textbf{1.94} & \textbf{1.79} \\
\cmidrule(lr){1-12}
\multirow{2}{*}{MT-Bench} & DFlash & 3.15 & 1.77 & 0.88 & 1.51 & 1.47 & 3.1 & 1.78 & 0.86 & 1.47 & 1.43 \\
 & DStill & 3.59 & \textbf{1.94} & \textbf{0.96} & \textbf{1.7} & \textbf{1.68} & 3.45 & \textbf{1.93} & \textbf{0.95} & \textbf{1.62} & \textbf{1.58} \\
\cmidrule(lr){1-12}
\multirow{2}{*}{SWE-Bench} & DFlash & 2.9 & 1.62 & 0.79 & 1.35 & 1.3 & 2.86 & 1.64 & 0.79 & 1.3 & 1.23 \\
 & DStill & 3.03 & \textbf{1.66} & \textbf{0.82} & \textbf{1.4} & \textbf{1.35} & 2.97 & \textbf{1.67} & \textbf{0.81} & \textbf{1.33} & \textbf{1.3} \\
\midrule
\multicolumn{12}{c}{\textit{Reasoning $=$ OFF}} \\
\midrule
\multirow{2}{*}{GSM8K} & DFlash & 5.65 & 3.1 & 2.27 & 3.33 & 2.22 & 4.88 & 2.71 & 1.85 & 2.69 & 2.01 \\
 & DStill & 7.02 & \textbf{3.75} & \textbf{2.66} & \textbf{3.83} & \textbf{2.47} & 5.9 & \textbf{3.37} & \textbf{2.13} & \textbf{3.04} & \textbf{2.24} \\
\cmidrule(lr){1-12}
\multirow{2}{*}{MATH-500} & DFlash & 4.95 & 2.69 & 1.4 & 2.47 & 2.09 & 4.43 & 2.54 & 1.27 & 2.23 & 1.96 \\
 & DStill & 5.51 & \textbf{2.91} & \textbf{1.56} & \textbf{2.77} & \textbf{2.31} & 4.92 & \textbf{2.68} & \textbf{1.44} & \textbf{2.42} & \textbf{2.19} \\
\cmidrule(lr){1-12}
\multirow{2}{*}{HumanEval} & DFlash & 5.99 & 3.2 & 2.15 & 3.07 & 2 & 5.32 & 2.88 & 1.89 & 2.74 & 1.83 \\
 & DStill & 7.46 & \textbf{3.88} & \textbf{2.64} & \textbf{3.43} & \textbf{2.28} & 6.52 & \textbf{3.46} & \textbf{2.26} & \textbf{3.22} & \textbf{1.95} \\
\cmidrule(lr){1-12}
\multirow{2}{*}{MBPP} & DFlash & 4.42 & 2.42 & 1.21 & 2.02 & 1.65 & 4.2 & 2.35 & 1.18 & 1.92 & 1.56 \\
 & DStill & 5.33 & \textbf{2.82} & \textbf{1.49} & \textbf{2.4} & \textbf{1.95} & 5.05 & \textbf{2.77} & \textbf{1.41} & \textbf{2.24} & \textbf{1.85} \\
\cmidrule(lr){1-12}
\multirow{2}{*}{MT-Bench} & DFlash & 2.91 & 1.7 & 0.89 & 1.45 & 1.42 & 2.76 & 1.61 & 0.84 & 1.36 & 1.21 \\
 & DStill & 3.3 & \textbf{1.91} & \textbf{0.98} & \textbf{1.67} & \textbf{1.5} & 3.13 & \textbf{1.78} & \textbf{0.94} & \textbf{1.52} & \textbf{1.4} \\
\cmidrule(lr){1-12}
\multirow{2}{*}{SWE-Bench} & DFlash & 2.98 & 1.68 & 0.88 & 1.46 & 1.35 & 2.83 & 1.65 & 0.84 & 1.36 & 1.24 \\
 & DStill & 3.22 & \textbf{1.77} & \textbf{0.93} & \textbf{1.54} & \textbf{1.45} & 3.11 & \textbf{1.77} & \textbf{0.87} & \textbf{1.51} & \textbf{1.32} \\
\bottomrule
\end{tabular}
}

\label{tab:nemotron_concurrency}
\end{table}

\subsection{Experimental Setup}
\label{sec:exp-setup}

\paragraph{Models and benchmarks.} We evaluate GPT-OSS-120B \citep{openai2025gptoss}, Gemma-4-31B
and Gemma-4-26B-A4B \citep{gemmateam2026gemma4}, and Nemotron-3.5-30B-A3B
\citep{nvidia2026nemotron35}, covering dense, transformer mixture-of-experts (MoE), and hybrid Mamba--attention MoE
architectures. We run the benchmarks on GSM8K \citep{cobbe2021training}, MATH-500
\citep{hendrycks2021measuring,lightman2024lets}, HumanEval \citep{chen2021evaluating}, MBPP
\citep{austin2021program}, SWE-bench \citep{jimenez2024swe}, and MT-Bench
\citep{zheng2023judging}. We evaluate 128 prompts each from
GSM8K, MATH-500, and MBPP; 164 each from HumanEval and SWE-bench;
and 80 from MT-Bench. Further evaluation details are provided
in Appendix~\ref{sec:exp-details}.

\paragraph{Training.} Each target regenerates assistant responses in OpenPerfectBlend
\citep{labonneopenperfectblend}; its hidden-state cache is shared across both training stages
and trajectory collection. Our \DStill drafters use block size
$\block=16$, comprising one anchor and 15 proposed tokens.
We collect teacher trajectories using $K=8$ refinement passes
with greedy token selection. Unless otherwise stated, student
distillation uses $\alpha=0.2$, top-$k=16$, and a cosine
root-probability schedule from $0.5$ to $0.85$.
The student is initialized from the teacher, and the target
embeddings and LM head remain frozen throughout training.
Drafter architectures are listed in Appendix
Table~\ref{tab:draft_configs_summary}, with further training
details in Appendix~\ref{sec:exp-details}.

\paragraph{Comparisons.} Our primary baseline is a draft model trained with the DFlash objective
\citep{chen2026dflash}. Within each target, this baseline and
\DStill share the drafter architecture, training corpus,
block size, and one-pass drafting budget.
For GPT-OSS-120B and Nemotron-3.5-30B-A3B, we evaluate the anchor-only
baseline and partial-mask teacher with one and eight passes
to measure the effect of iterative refinement and its transfer
to the student (Table~\ref{tab:main-results}).
We also apply trajectory distillation to a DSpark drafter
(Appendix Table~\ref{tab:dsparkplusdstill}).
Autoregressive target decoding provides the speedup reference.

\paragraph{Inference setup.}
Table~\ref{tab:model_performance} reports results at request
concurrency $C=1$ and temperature 0, with reasoning enabled
and reasoning effort set to \texttt{medium} for GPT-OSS-120B.
For GPT-OSS-120B and Nemotron-3.5-30B-A3B, we additionally evaluate
$C\in\{1,8,16,32\}$ at temperatures 0 and 1; Nemotron is
evaluated with reasoning both enabled and disabled
(Tables~\ref{tab:gptoss_concurrency}
and~\ref{tab:nemotron_concurrency}).
Prompts, generation settings, and hardware are matched across
methods within each comparison.

\paragraph{Evaluation metrics.} Mean acceptance length $\tau$
is the average number of tokens in the accepted block prefix,
including the anchor token.
We report speculative decoding speedup over autoregressive
target decoding as $S=t_{\mathrm{AR}}/t_{\mathrm{spec}}$,
where $t_{\mathrm{AR}}$ and $t_{\mathrm{spec}}$ denote generation
time per output token under autoregressive and speculative
decoding, respectively. Averages reported in the tables are unweighted means across benchmarks or concurrency levels, as indicated. \kbcomment{is this the precise definition of the metric ? If this is true shouldnt Speedup and Tau be perfectly correlated ? What explains the fact that its not ? Prefill time ? but then this definition is imprecise}

\subsection{Main Results}
\label{sec:exp-main}

\paragraph{Benchmark comparison.} We compare \DStill
with \DFlash on reasoning, mathematics, code generation, and
conversation benchmarks for three model families across temperature and reasoning settings in Tables~\ref{tab:model_performance},~\ref{tab:gptoss_concurrency}, and ~\ref{tab:nemotron_concurrency}. We train iso-parameter models for both the methods using the OpenPerfectBlend dataset.

\paragraph{Performance across reasoning modes.}
With reasoning enabled, \DStill consistently outperforms \DFlash by $6.7\%$-$18.8\%$ on average speedups over autoregressive decoding and $7.6\%$-$21.1\%$ on acceptance length  as shown in Table~\ref{tab:model_performance}. On GPT-OSS-120B, \DStill shows speed up improvements over \DFlash across temperature and concurrency settings as in Table~\ref{tab:gptoss_concurrency}. 
In non-thinking mode, Nemotron achieves speedups of up to $3.88\times$ and $3.46\times$ at temperatures 0 and 1, respectively, at concurrency 1
(Table~\ref{tab:nemotron_concurrency}).

\paragraph{Transfer of refinement gains.}
Table~\ref{tab:main-results} compares \DFlash, the teacher,
and \DStill across refinement steps.
Eight teacher passes improve acceptance length by 18.0\%
on GPT-OSS-120B and 17.9\% on Nemotron.
One-pass \DStill recovers 35.8\% and 28.6\% of these gains,
outperforming \DFlash by 7.6\% and 19.4\%, respectively.
The largest teacher--student gap is on Nemotron HumanEval
(8.57 vs.\ 7.46; 1.11 tokens).
Eight-pass \DFlash even underperforms the one-pass teacher,
likely due to a mismatch between its root-only training
and partially revealed refinement states, motivating Stage~I.

\begin{table*}[t]
  \centering
  \caption{Draft acceptance across target models and benchmarks.
  Denoising steps denote sequential drafter forward passes per block.
  Average is the unweighted mean across the six benchmarks.
  Recovery is the percentage of the teacher's one-to-eight-pass
  gain in average acceptance length recovered by one-pass \DStill. Thinking mode is \textit{disabled} in the measurements with Nemotron-3.5-30B-A3B so the numbers will be different from Table~\ref{tab:model_performance}}
  \label{tab:main-results}
  \scriptsize
  \setlength{\tabcolsep}{3pt}
  \begin{tabular}{@{}llc*{8}{c}@{}}
    \toprule
    \multirow{2}{*}{\textbf{Target Model}} &
    \multirow{2}{*}{\textbf{Drafter}} &
    \multirow{2}{*}{\shortstack{\textbf{Denoising}\\\textbf{Steps}}} &
    \textbf{GSM8K} & \textbf{MATH-500} &
    \textbf{HumanEval} & \textbf{MBPP} &
    \textbf{MT-Bench} & \textbf{SWE-Bench} &
    \textbf{Average} &
    \multirow{2}{*}{\shortstack{\textbf{Recovery}\\\textbf{(\%)}}} \\
    \cmidrule(lr){4-4}
    \cmidrule(lr){5-5}
    \cmidrule(lr){6-6}
    \cmidrule(lr){7-7}
    \cmidrule(lr){8-8}
    \cmidrule(lr){9-9}
    \cmidrule(lr){10-10}
    & & &
    $\tau$ & $\tau$ & $\tau$ & $\tau$ &
    $\tau$ & $\tau$ & $\tau$ & \\
    \midrule

    \multirow{5}{*}{GPT-OSS-120B}
      & DFlash & 1 & 6.30 & 5.17 & 4.58 & 4.63 & 3.09 & 2.96 & 4.45 & -- \\
      & DFlash & 8 & 4.97 & 4.06 & 3.75 & 3.76 & 2.88 & 2.79 & 3.70 & -- \\
      & Teacher & 1 & 6.44 & 5.10 & 4.63 & 4.65 & 3.19 & 2.97 & 4.50 & -- \\
      & Teacher & 8 & 7.58 & 6.14 & 5.48 & 5.62 & 3.60 & 3.41 & 5.31 & -- \\
      & \DStill & 1 & 6.79 & 5.54 & 4.83 & 4.97 & 3.34 & 3.25 & 4.79 & 35.80 \\
    \midrule

    \multirow{5}{*}{\shortstack[l]{Nemotron-3.5-\\30B-A3B}}
      & DFlash & 1 & 5.84 & 4.84 & 5.98 & 4.44 & 2.90 & 2.92 & 4.49 & -- \\
      & DFlash & 8 & 4.94 & 4.19 & 4.70 & 3.90 & 2.84 & 2.78 & 3.89 & -- \\
      & Teacher & 1 & 6.77 & 5.43 & 6.85 & 5.02 & 3.18 & 3.31 & 5.10 & -- \\
      & Teacher & 8 & 7.80 & 6.48 & 8.57 & 6.17 & 3.51 & 3.52 & 6.01 & -- \\
      & \DStill& 1 & 6.94 & 5.65 & 7.46 & 5.40 & 3.32 & 3.35 & 5.36 & 28.57 \\

    \bottomrule
  \end{tabular}
  
\end{table*}

% \paragraph{Decoding speed.} The latency measurements assess the cost of each drafting
% configuration. \DStill achieves \textcolor{red}{XX}$\times$--\textcolor{red}{XX}$\times$
% speedup over autoregressive decoding and changes per-token latency by \textcolor{red}{XX}\%
% relative to the anchor-only drafter and \textcolor{red}{XX}\% relative to the eight-pass teacher.
% \textcolor{red}{The acceptance-length changes translate into XX across targets; the latency
% breakdown attributes the difference to XX.}

\paragraph{Serving performance.}
\DStill improves throughput across all reported settings,
achieving speedups of up to $5.2\times$ for GPT-OSS-120B
and $3.0\times$ for Nemotron-3.5-30B-A3B over
autoregressive decoding.
At concurrency 32, speedups remain $2.7$--$3.9\times$
and $2.1$--$2.4\times$, respectively, showing that the
decoding gains persist under concurrent serving.

\subsection{Ablation Studies}
\label{sec:exp-ablations}

We examine the effects of continued baseline training and
student state sampling. Evaluations use mean acceptance
length with greedy decoding and one drafting pass on
$8\times$H100 GPUs. Unless otherwise stated, student
distillation runs for 3 epochs with $\alpha=0.2$ and a
cosine $p_{\mathrm{root}}$ schedule from $0.5$ to $0.85$.
Additional ablations are reported in
Appendix~\ref{sec:extra-ablations}.

\paragraph{Continued one-step training.} Table~\ref{tab:DFlash-training-duration} compares \DFlash trained for 10 epochs with \DStill in its default setting on Gemma-4-31B. \DStill achieves higher acceptance length across all six benchmarks, with improvements ranging from 0.21 tokens on SWE-bench to 0.7 tokens on GSM8K. These results suggest that trajectory distillation
provides gains beyond longer \DFlash training.

\begin{table}[H]
  \caption{Acceptance length of \DStill and \DFlash trained for 10 epochs on Gemma-4-31B.}
  \label{tab:DFlash-training-duration}
  \centering
  \small
  \begin{tabular}{lrrr}
    \toprule
    \textbf{Dataset} &
    \shortstack{\textbf{\DFlash}\\\textbf{(5 epochs)}$\uparrow$} &
    \shortstack{\textbf{\DFlash}\\\textbf{(10 epochs)}$\uparrow$} &
    \shortstack{\textbf{\DStill}\\\textbf{(default)}$\uparrow$} \\
    \midrule
    GSM8K     & 6.78 & 6.54 & \textbf{7.48} \\
    MATH500 & 6.84 & 6.67 & \textbf{7.28} \\
    HumanEval   & 6.03 & 5.96 & \textbf{6.49} \\
    MBPP      & 5.00 & 5.00 & \textbf{5.52} \\
    MT-Bench  & 3.67 & 3.57 & \textbf{4.03} \\
    SWE-Bench & 3.17 & 3.13 & \textbf{3.38} \\
    \bottomrule
  \end{tabular}
\end{table}

\paragraph{State sampling.} We compare root-only training, which samples exclusively from the fully masked block state ($p_{\mathrm{root}}=1$), with our cosine schedule, which progressively increases the probability of sampling this state from $0.5$ to $0.85$. We trained the drafters for 16 epochs, and conduct this ablation on one-eighth of the training dataset,
with all other training settings held fixed. Table~\ref{tab:state-sampling} shows the cosine schedule improves $\tau$ across all six benchmarks, with the largest gains on GSM8K ($+0.22$) and MBPP ($+0.10$). These results suggest that progressively increasing the probability of sampling the fully masked state, while retaining exposure to intermediate trajectory states, improves student acceptance length.

\begin{table}[H]
  \caption{Mean acceptance length $\tau$ for root-only
and cosine state sampling on Gemma-4-26B-A4B
at concurrency 16.}
\label{tab:state-sampling}
  \centering
  \small
  \begin{tabular}{lrrr}
    \toprule
    \textbf{Dataset} &
    $\boldsymbol{p_{\mathrm{root}}=1}$$\uparrow$ &
    \textbf{Cosine}$\uparrow$ & $\boldsymbol{\Delta}$ \\
    \midrule
    GSM8K     & 4.77 & \textbf{4.99} & $+0.22$ \\
    HumanEval & 4.58 & \textbf{4.60} & $+0.02$ \\
    MATH500   & 4.90 & \textbf{4.97} & $+0.07$  \\
    MBPP      & 4.11 & \textbf{4.21} & $+0.10$  \\
    MT-Bench  & 3.16 & \textbf{3.19} & $+0.03$  \\
    SWE-Bench & 2.57 & \textbf{2.58} & $+0.01$  \\
    \midrule
    Average & 4.01 & \textbf{4.09} & $+0.08$ \\
    \bottomrule
  \end{tabular}
\end{table}

\section{Conclusion}
\label{sec:conclusion}

We presented a trajectory consistency distillation procedure for block-diffusion speculative decoding: a two-stage training
pipeline that trains a multi-step masked-diffusion drafter with
partial-mask supervision and then distills its confidence-guided denoising trajectory into a one-step student via an offline trajectory-replay pipeline. Distillation trains the student to map any
state along the teacher's trajectory, including the root state used at inference time, directly to the teacher's endpoint, under a curriculum that shifts weight toward the root state as training progresses. We show that this procedure recovers $28.6\%$–$35.8\%$ of the acceptance length improvements of a multi-step teacher in a single step student with zero test-time overheads. Across models and domains, we demonstrate improvements of $28\%$ in acceptance length and $25\%$ in speedup. Finally, we demonstrate our procedure's general applicability to diffusion speculators by improving DSpark acceptance lengths by $4.5\%$. 

% \clearpage
% Restore the full column height after clearing pending two-column floats.
\makeatletter
\global\@colroom=\@colht
\global\vsize=\@colht
\makeatother
% Avoid stretched vertical gaps in the references and appendix.
\raggedbottom
\section*{Limitations}

\DStill has a higher one-time training cost than direct \DFlash training, comprising
partial-mask teacher training, offline trajectory collection and storage, and student distillation.
These additional costs can be amortized over subsequent generation, but only at a large enough inference workload.
The student learns the teacher's selected continuations and inherits errors in
those endpoints, and the student's one-pass outputs also remain factorized. Our experiments only cover the target
models, drafter configurations, and benchmarks reported here. Transfer to other block sizes,
architectures beyond \DFlash and DSpark, and target-feature layouts requires further evaluation. Acceptance-length gains
must be considered together with measured latency under the inference workload. Actual latency reductions depend on the specific inference workload, hardware serving kernels, and concurrency limits.

\bibliography{references}

\appendix
\section{Drafting and Verification Details}
\label{sec:prelim-details}

\subsection{Drafting Setup}

Let $\mathcal T$ denote the target language model and $(u_{1:S},a)$ the current verified prefix,
where $a$ is its most recent token. We cache the target's hidden states $\Hc$ at a fixed set of
layers over $u_{1:S}$. A draft block contains $\block$ positions: position $0$ holds the fixed
anchor $a$, and positions $1,\ldots,\block-1$ are the future tokens to be proposed. Uncommitted
positions contain $\mask$, so the initial block is $x_1=(a,\mask,\ldots,\mask)$.

The drafter $D_\theta$ receives embeddings of the current block state $x$ together with $\Hc$,
attends bidirectionally within the block and to context strictly before its anchor, and produces
a vocabulary distribution $p_{\theta,i}(\cdot\mid x,\Hc)$ at each masked position $i$.
Independently sampling all non-anchor positions from the root predictions gives
the factorized proposal in Equation~\ref{eq:one-step-proposal}.
Greedy drafting selects the most probable token at each position. In both cases, all positions
are predicted from the same root state, yielding the one-pass factorization discussed in
Section~\ref{sec:intro}.

The target scores the proposal in one causal forward pass. Standard lossless speculative
sampling uses the target's prefix-conditional probabilities and the proposal's actual
left-to-right conditionals, together with rejection correction
\citep{leviathan2023fast,chen2023accelerating}. Under greedy verification, acceptance is determined
by matching the target's top-1 predictions. Let $L\in\{0,\ldots,\block-1\}$ be the length of the
accepted draft prefix.
We denote mean \emph{acceptance length} by $\tau=1+\mathbb E[L]$.

\subsection{Denoising Trajectories}

A multi-step drafter resolves the block over $K$ sequential passes. Let $x_t\equiv x^{(t)}$
denote the state before pass $t\in\{1,\ldots,K\}$, starting from the anchor-only root $x_1$.
The set of positions already committed is $c^{(t)}\subseteq\{0,\ldots,\block-1\}$, with
$c^{(1)}=\{0\}$ and cardinality $|c^{(t)}|$. Thus the root contains the given anchor but
zero committed \emph{future} tokens. The remaining masked positions are
$M_t=\{1,\ldots,\block-1\}\setminus c^{(t)}$; we write $M$ when the pass index is unnecessary.

At pass $t$, the drafter predicts from $(x_t,\Hc)$ and assigns each masked position confidence
$\text{conf}_i=\max_v p_{\theta,i}(v\mid x_t,\Hc)$. It commits the
\begin{equation}
  \label{eq:reveal-schedule}
  n_t \;=\; \left\lceil \frac{\block-|c^{(t)}|}{K-t+1} \right\rceil
\end{equation}
highest-confidence positions, writing their greedy or sampled tokens into the block and leaving
earlier commitments unchanged. The updated state $x^{(t+1)}$ is the input to the next pass.
The schedule allocates the remaining masked positions across the remaining passes, committing
all positions after $K$ passes. The rollout starts from the root predictions in
Equation~\ref{eq:one-step-proposal}; later passes condition on tokens committed earlier.

We reserve $x_0\equiv x^{(K+1)}=(a,y_1,\ldots,y_{\block-1})$ for the fully committed teacher
endpoint. Here, $t$ indexes forward passes and the subscript $0$ identifies the completed block. The
student's proposed block is $\hat x_0=(a,\hat y_1,\ldots,\hat y_{\block-1})$.
Figure~\ref{fig:dstill-overview} uses these state conventions. Lossless verification of a sampled
multi-step proposal requires its actual left-to-right conditionals, which can differ from the
distributions recorded at commitment time.

This rollout defines the intermediate states $x_t$ and endpoint $x_0$ used in
Section~\ref{sec:method}: Stage I trains the teacher across intermediate corruption levels, and
Stage II uses recorded teacher trajectories to supervise a one-pass student.

\section{Training and Trajectory Details}
\label{sec:training-details}

\subsection{Reveal schedule and teacher objective}
\label{sec:teacher-details}

During training, we cache $\Hc$ from teacher-forced sequences and construct blocks around
sampled anchor positions. Each block's attention is restricted to its own positions and target
context strictly before its anchor, with bidirectional attention within the block. This attention mask
prevents cross-block leakage and allows multiple anchored blocks to be scored in parallel.
Let $z=(a,z_1,\ldots,z_{\block-1})$ denote a reference training block.

We derive corruption levels from the number of positions committed before each pass of
Equation~\ref{eq:reveal-schedule}. Let $m_t$ include the anchor. The corresponding reveal
probabilities for \emph{non-anchor} positions are
\begin{equation}
  \label{eq:reveal-levels}
  \begin{aligned}
    m_1 &= 1, \\
    m_{t+1} &= m_t + \left\lceil \frac{\block-m_t}{K-t+1} \right\rceil, \\
    \mathcal R &= \left\{\frac{m_t-1}{\block-1}:
      1\le t\le K,\ m_t<\block\right\}.
  \end{aligned}
\end{equation}
For each block, we sample $r$ uniformly from $\mathcal R$ and independently reveal each
non-anchor position with probability $r$; the anchor is always revealed. Revealed positions use
their reference-token embeddings, and the others use $\mask$, yielding the corrupted block $x^r$.
The reveal level and Bernoulli mask are sampled anew when a block is revisited across epochs.
The root level is $r=0$. Each level matches the expected number of revealed positions to a
rollout state. Bernoulli sampling varies the revealed counts and position subsets. Fully committed levels are excluded
from $\mathcal R$. Any Bernoulli outcome with no masked positions contributes no loss and is
excluded from the training average.

For a nonempty masked set $M$, the per-block objective is
\begin{equation}
  \label{eq:partial-mask-loss}
  \begin{aligned}
    \ell_i^{\mathrm{CE}} &= -\log p_{\theta,i}(z_i\mid x^r,\Hc), \\
    \mathcal L_{\mathrm{teacher}}
      &= w_{\mathrm{block}}\,
      \frac{\sum_{i\in M} w_{\mathrm{pos}}(i)\,\ell_i^{\mathrm{CE}}}
           {\sum_{i\in M} w_{\mathrm{pos}}(i)}.
  \end{aligned}
\end{equation}
Training minimizes the arithmetic mean of this objective over nonempty blocks; batches with no
nonempty blocks are skipped. The block weight scales each block's normalized position loss and
controls its contribution to the training average. Both weights are optional and default to $1$.

\paragraph{Mask-count weighting.} To emphasize heavily masked blocks relative to nearly complete
ones, we use
\begin{equation}
  \label{eq:mask-count-weight}
  w_{\mathrm{block}} =
  \frac{(|M|/(\block-1))^{\alpha_m}}
       {\mathbb E_{j\sim\mathrm{Unif}\{1,\ldots,\block-1\}}
       [(j/(\block-1))^{\alpha_m}]}.
\end{equation}
Setting $\alpha_m=0$ gives $w_{\mathrm{block}}=1$, while $\alpha_m>0$ increases the relative
weight of heavily masked blocks. The denominator normalizes against a uniform \emph{reference}
distribution over mask counts. The mean weight under the training sampler depends on its
mask-count distribution.

\paragraph{Position decay.} Following DFlash \citep{chen2026dflash}, we optionally give greater
weight to early positions because acceptance depends on a prefix. We use
$w_{\mathrm{pos}}(i)=\exp(-(i-1)/\gamma)$ for $\gamma>0$.
With position decay disabled, $w_{\mathrm{pos}}(i)=1$.

The discrete levels emphasize corruption depths associated with the teacher's refinement
schedule, while random reveal subsets broaden the training distribution. This aligns expected
reveal counts; the positions and reference tokens differ from the confidence-selected
predictions exposed during a teacher rollout.
Stage-I examples reveal \emph{reference tokens} independently; they are not connected rollouts
through the drafter's own predictions. They therefore do not supply the trajectory endpoints
needed for distillation. We obtain these by running the frozen Stage-I checkpoint through the
full refinement schedule. Equivalent trajectories already recorded from that checkpoint under
the same rollout policy could be reused; no additional teacher-training stage is needed.

\subsection{Trajectory records and reconstruction}
\label{sec:trajectory-details}

We run the frozen teacher's confidence-guided rollout offline, once per training block, and
record the tokens it commits. Each record contains
\begin{itemize}
  \item $x_0\in\{1,\ldots,V\}^{\block}$: the fully committed endpoint, where $V$ is the vocabulary size;
  \item $\kappa\in\{0,\ldots,K\}^{\block}$: each position's commit pass, with $\kappa_0=0$;
  \item $(\mathrm{ids}^{(t)}_i,\mathrm{logp}^{(t)}_i)$: the top-$k$ token ids and their
  full-softmax log-probabilities at each position while it remains masked.
\end{itemize}
Because commitments are never revised, $x_0$ and $\kappa$ reconstruct every intermediate block
without storing a separate $K\times\block$ sequence of block states:
\begin{equation}
  \label{eq:trajectory-reconstruction}
  x_{t,i} =
  \begin{cases}
    x_{0,i}, & \kappa_i<t, \\
    \mask, & \kappa_i\ge t.
  \end{cases}
\end{equation}
For example, a token committed at pass 2 is masked in the input to that pass and visible in
the next state.
The revealed token is the \emph{teacher-committed} token $x_{0,i}$, which may differ from the
reference token $z_i$.

For each non-anchor position $i$, let $S_i=\mathrm{ids}^{(\kappa_i)}_i$ be the retained ids at its
committing pass. For $v\in S_i$, recover the original teacher probability as
$q_{i,v}=\exp(\mathrm{logp}^{(\kappa_i)}_i(v))$, where the stored log-probability corresponds to
id $v$. These probabilities retain their full-vocabulary normalization. Define $q_{i,\mathrm{other}}=1-\sum_{v\in S_i}q_{i,v}$ and let
$Q_{\mathrm{end},i}$ denote the resulting distribution over $S_i\cup\{\mathrm{other}\}$.
We write $Q_{\mathrm{end}}=(Q_{\mathrm{end},i})_{i=1}^{\block-1}$. Thus, the record supports
both exact state reconstruction and committing-pass distribution supervision.

\begin{algorithm}[t]
  \caption{Offline Trajectory Collection}
  \label{alg:collect}
  \small
  \raggedright
  \renewcommand{\algorithmicrequire}{\textbf{Input:}}
  \renewcommand{\algorithmicensure}{\textbf{Output:}}
  \begin{algorithmic}[1]
    \REQUIRE Frozen teacher $D_\theta$, context $\Hc$, anchor $a$, block size $\block$, steps $K$, top-$k$
    \ENSURE Endpoint $x_0$, commit passes $\kappa$, top-$k$ records
    \STATE $x^{(1)}\gets(a,\mask,\ldots,\mask)$
    \STATE $c^{(1)}\gets\{0\}$; $\kappa_0\gets0$
    \FOR{$t=1$ \TO $K$}
      \STATE $M_t\gets\{1,\ldots,\block-1\}\setminus c^{(t)}$
      \STATE $\mathrm{logits}\gets D_\theta(x^{(t)},\Hc)$
      \STATE $P^{(t)}_i\gets\mathrm{softmax}(\mathrm{logits}_i)$, $i\in M_t$
      \STATE $\mathrm{conf}_i\gets\max_v P^{(t)}_{i,v}$, $i\in M_t$
      \STATE Store top-$k$ ids in $\mathrm{ids}^{(t)}$
      \STATE Store their $\log P^{(t)}_{i,v}$ in $\mathrm{logp}^{(t)}$
      \STATE $n_t\gets\lceil(\block-|c^{(t)}|)/(K-t+1)\rceil$
      \STATE $J_t\gets$ top $n_t$ indices in $M_t$ by confidence
      \STATE $x^{(t+1)}\gets x^{(t)}$
      \STATE Fill $J_t$ with greedy or sampled teacher tokens
      \STATE $\kappa_i\gets t$, $i\in J_t$
      \STATE $c^{(t+1)}\gets c^{(t)}\cup J_t$
    \ENDFOR
    \STATE $x_0\gets x^{(K+1)}$
    \RETURN $x_0,\kappa,\{(\mathrm{ids}^{(t)},\mathrm{logp}^{(t)})\}_{t=1}^K$
  \end{algorithmic}
\end{algorithm}

\subsection{State-sampling schedule}
\label{sec:root-details}
 With probability $p_{\mathrm{root}}$, we sample the
anchor-only root $x_1$ ($t=1$). Otherwise, we sample $t$ uniformly from
$\{2,\ldots,\kappa_{\max}\}$, where $\kappa_{\max}=\max_i\kappa_i$ is the last commit pass.
Blocks with no intermediate state fall back to $t=1$. We resample $t$ on each visit to a
trajectory across epochs, reconstructing the corresponding recorded state each time.
The endpoint $x_0$ and committing-pass distributions remain fixed. Thus, different partial
commitments provide supervision toward the same completed block; the student learns to
predict all remaining scored positions in one pass. This combines the root state with
intermediate states that provide additional endpoint supervision. Teacher-committed tokens in
intermediate states provide additional context for predicting the remaining masked positions
early in training. We ramp the root probability with training progress $\rho\in[0,1]$:
\begin{equation}
  \label{eq:root-schedule}
  \begin{aligned}
    u(\rho) &= \frac{\rho-w}{1-c-w}, \\
    s(\rho) &= \frac{1-\cos(\pi u(\rho))}{2}, \\
    \Delta p &= p_{\mathrm{end}}-p_{\mathrm{start}}, \\
    p_{\mathrm{root}}(\rho) &=
    \begin{cases}
      p_{\mathrm{start}}, & \rho\le w, \\
      p_{\mathrm{start}}+\Delta p\,s(\rho), & w<\rho<1-c, \\
      p_{\mathrm{end}}, & \rho\ge1-c.
    \end{cases}
  \end{aligned}
\end{equation}
Here $w,c\ge0$ are warmup and cooldown fractions with $w+c<1$. We use
$p_{\mathrm{start}}=0.5$, $p_{\mathrm{end}}=0.85$, $w=0.10$, and $c=0.30$.
The schedule gives intermediate states more weight early and increasingly emphasizes the
root state. In particular, the root receives direct supervision throughout
training, with sampling probability $0.5$ initially and $0.85$ late in training. At the root,
every scored non-anchor position is masked and receives endpoint supervision. Root behavior
is therefore trained explicitly rather than inferred only from partially committed inputs.

\subsection{Distillation distributions and loss aggregation}
\label{sec:loss-details}
 We reconstruct $x_t$ using Equation~\ref{eq:trajectory-reconstruction} and
restrict $M_t$ to still-masked, scored response positions, excluding the anchor.
Scored positions are response-token positions selected for supervision by the training loss mask.
For $v\in S_i$, let $p_{i,v}=p_{\phi,i}(v\mid x_t,\Hc)$ and
$p_{i,\mathrm{other}}=1-\sum_{v\in S_i}p_{i,v}$. Define $P_{\phi,i}$ over the same
$k+1$ categories as $Q_{\mathrm{end},i}$. The per-position objective is Equation~\ref{eq:distill-loss}.
The CE term targets the teacher's committed token, while KD transfers its committing-pass
probability mass, including the omitted tail:
\begin{equation}
  \label{eq:kd-with-other}
  \begin{aligned}
    D_{\mathrm{KL}}(Q_{\mathrm{end},i}\,\|\,P_{\phi,i})
      &=\sum_{v\in S_i}q_{i,v}\log\frac{q_{i,v}}{p_{i,v}} \\
      &\quad+q_{i,\mathrm{other}}\log
        \frac{q_{i,\mathrm{other}}}{p_{i,\mathrm{other}}}.
  \end{aligned}
\end{equation}
The KL divergence is computed over the retained token ids and a single aggregate tail category.
Storing the teacher's top-$k$ probabilities is sufficient to evaluate this coarsened loss.
We use $\alpha=0.2$; Section~\ref{sec:experiments} reports its ablation.

\paragraph{Trajectory-quality weighting.} Optionally, we favor positions where the teacher's
endpoint agrees with the reference training block $z$. Let $d$ be the first scored non-anchor
position with $x_{0,d}\ne z_d$, or $d=\block$ if there is no disagreement. We use
\begin{equation}
  \label{eq:traj-weight}
  w_j=
  \begin{cases}
    1, & j<d, \\
    w_{\mathrm{first}}, & j=d, \\
    w_{\mathrm{first}}\,\beta^{j-d}, & j>d,
  \end{cases}
\end{equation}
with $0<w_{\mathrm{first}}\le1$ and $0<\beta<1$; a flat tail weight applies if $\beta$ is unset.
The weights provide a \emph{reference-agreement heuristic} that reduces supervision after a
mismatch while retaining the trajectory. A teacher token that differs from a sampled reference
continuation can still be accepted under lossless speculative sampling.

For a batch of blocks indexed by $b$, we aggregate over scored masked positions using
\begin{equation}
  \label{eq:student-loss}
  \mathcal L_{\mathrm{student}} =
  \frac{\sum_b\sum_{i\in M_{t_b}}w_{b,i}\,\mathcal L_{b,i}}
       {\sum_b\sum_{i\in M_{t_b}}w_{b,i}}.
\end{equation}
When trajectory weighting is disabled, $w_{b,i}=1$, recovering the ordinary mean token loss.
Only blocks with scored masked positions enter the batch loss; batches with no such positions
are skipped. Algorithm~\ref{alg:distill} summarizes one update.

\begin{algorithm}[t]
  \caption{Trajectory Consistency Distillation}
  \label{alg:distill}
  \small
  \raggedright
  \renewcommand{\algorithmicrequire}{\textbf{Input:}}
  \renewcommand{\algorithmicensure}{\textbf{Output:}}
  \begin{algorithmic}[1]
    \REQUIRE Student $D_\phi$, training batch with stored trajectories, contexts $\Hc$, progress $\rho$
    \ENSURE Updated $\phi$, unless the batch has zero weight
    \STATE $p_{\mathrm{root}}\gets$ Equation~\ref{eq:root-schedule} at $\rho$
    \STATE $\mathrm{lossSum}\gets0$; $\mathrm{weightSum}\gets0$
    \FOR{each block in the batch}
      \STATE Resample $t$ from the root-biased schedule
      \STATE Reconstruct $x_t$ using Equation~\ref{eq:trajectory-reconstruction}
      \STATE $M_t\gets\{i:\kappa_i\ge t,\ i\text{ scored},\ i\ne0\}$
      \IF{$M_t\ne\emptyset$}
        \STATE $\mathrm{logits}\gets D_\phi(x_t,\Hc)$
        \STATE Recover $Q_{\mathrm{end},i}$ at pass $\kappa_i$, $i\in M_t$
        \STATE Compute $\mathcal L_i$ using Equation~\ref{eq:distill-loss}, $i\in M_t$
        \STATE Set $w_i$ using Equation~\ref{eq:traj-weight}, or $1$ if disabled
        \STATE $\mathrm{lossSum}\gets\mathrm{lossSum}+\sum_{i\in M_t}w_i\mathcal L_i$
        \STATE $\mathrm{weightSum}\gets\mathrm{weightSum}+\sum_{i\in M_t}w_i$
      \ENDIF
    \ENDFOR
    \IF{$\mathrm{weightSum}>0$}
      \STATE Backpropagate $\mathrm{lossSum}/\mathrm{weightSum}$
      \STATE Update student parameters $\phi$
    \ENDIF
  \end{algorithmic}
\end{algorithm}

\section{Ablation Results}
\label{sec:extra-ablations}

\paragraph{Student supervision.} To investigate the effect of the supervision objective, we compare endpoint CE ($\alpha=0$), KD ($\alpha=1$), and their combination, CE+KD ($\alpha=0.2$), in Equation~\ref{eq:distill-loss}. As shown in Table~\ref{tab:student-supervision}, the three objectives achieve broadly comparable acceptance lengths across benchmarks. CE and KD provide complementary supervision signals, combining endpoint-level targets with distributional information from the teacher.

\begin{table}[H]
  \caption{Student supervision ablation at concurrency 16.
  Higher $\tau$ is better. $\alpha=0$ denotes endpoint CE,
  $\alpha=0.2$ denotes CE+KD (ours), and $\alpha=1$ denotes KD.}
  \label{tab:student-supervision}
  \centering
  \small
  \begin{tabular}{lrrr}
    \toprule
    \textbf{Dataset} & $\boldsymbol{\alpha=0}$ &
    $\boldsymbol{\alpha=0.2}$ & $\boldsymbol{\alpha=1}$ \\
    \midrule
    GSM8K     & 5.6 & 5.4 & 5.5 \\
    HumanEval & 5.0 & 5.1 & 5.1 \\
    MATH500   & 5.7 & 5.6 & 5.6 \\
    MBPP      & 4.5 & 4.4 & 4.5 \\
    MT-Bench  & 3.3 & 3.4 & 3.3 \\
    SWE-Bench & 2.8 & 2.8 & 2.8 \\
    \bottomrule
  \end{tabular}
\end{table}

% \paragraph{Partial-mask teacher training.} We measure how training across reveal levels affects
% iterative refinement and its transfer to the student. Teachers use anchor-only masks, continuous
% reveal ratios $r\sim\mathrm{Unif}(0,1)$, or the discrete levels in Equation~\ref{eq:reveal-levels}.
% Each teacher supplies trajectories for a student trained with the same distillation configuration.
% \textcolor{red}{Partial-mask training changes teacher and student acceptance lengths by XX and
% XX tokens relative to anchor-only training. Discrete-level sampling changes them by XX and XX
% relative to continuous sampling.}

\paragraph{Trajectory-quality weighting.} The distillation loss can optionally down-weight
supervision at positions after the teacher's recorded endpoint first diverges from the reference
continuation, following the reference-agreement heuristic in Equation~\ref{eq:traj-weight}.
We compare a student distilled with uniform position weights against one that
enables this weighting with $w_{\mathrm{first}}=0.4$ and tail decay $\beta=0.5$.
Table~\ref{tab:traj-weight} reports student acceptance length. Weighting increases $\tau$ on all six benchmarks, by $0.04$ to $0.12$ tokens, raising
the average from $4.72$ to $4.80$ ($+1.7\%$). Concentrating supervision on the agreed prefix,
rather than on lower-agreement suffix positions that are less likely to be accepted, improves
one-step drafting at no additional inference cost.

\begin{table}[H]
  \caption{Trajectory-quality weighting ablation on GPT-OSS-120B. The weighted run additionally enables the reference-agreement
  weighting of Equation~\ref{eq:traj-weight} with $w_{\mathrm{first}}{=}0.4$ and tail decay
  $\beta{=}0.5$.}
  \label{tab:traj-weight}
  \centering
  \small
  \begin{tabular}{lrrr}
    \toprule
    \textbf{Dataset} & \textbf{Unweighted}$\uparrow$ & \textbf{Weighted}$\uparrow$ & $\boldsymbol{\Delta}$ \\
    \midrule
    GSM8K     & 6.96 & \textbf{7.01} & $+0.05$ \\
    MATH-500  & 5.35 & \textbf{5.47} & $+0.12$ \\
    HumanEval & 4.77 & \textbf{4.88} & $+0.11$ \\
    MBPP      & 4.95 & \textbf{4.99} & $+0.04$ \\
    MT-Bench  & 3.26 & \textbf{3.32} & $+0.06$ \\
    SWE-Bench & 3.06 & \textbf{3.15} & $+0.09$ \\
    \midrule
    Average   & 4.72 & \textbf{4.80} & $+0.08$ \\
    \bottomrule
  \end{tabular}
\end{table}

\section{Additional Experimental Results}
\label{sec:extra-results}

Table~\ref{tab:main-results} compares drafting cost and acceptance for the anchor-only baseline,
the teacher evaluated with one or eight passes, and the distilled one-pass student.
The teacher comparison isolates the refinement gain available for distillation.

\section{Experimental Details}
\label{sec:exp-details}

\paragraph{Verifier models.}
Table~\ref{tab:target-models} lists the target-model
specifications. The $A$ suffix denotes
nominal active parameters for the Gemma and Nemotron MoE models.

\begin{table*}[t]
  \caption{Target specifications from the model reports
\citep{openai2025gptoss,gemmateam2026gemma4,nvidia2026nemotron35}
and released configurations.}
  \label{tab:target-models}
  \begin{center}
  \begin{small}
  \resizebox{\linewidth}{!}{%
  \begin{tabular}{lccccc}
    \toprule
    \textbf{Target Model} & \textbf{Size} & \textbf{Architecture} & \textbf{Precision} & \textbf{Context} & \textbf{\# Layers} \\
    \midrule
    GPT-OSS-120B \citep{openai_gpt_oss_120b_2026} & 120B & MoE & BF16 & 131{,}072 (128K) & 36 \\
    Gemma-4-31B \citep{google_gemma_4_31b_it_2026} & 31B & Dense & BF16 & 262{,}144 (256K) & 60 \\
    Gemma-4-26B-A4B \citep{google_gemma_4_26b_a4b_it_2026} & 26B-A4B & MoE & BF16 & 262{,}144 (256K) & 30 \\
    Nemotron-3.5-30B-A3B \citep{nvidia2026nemotron35} & 30B-A3B & Hybrid MoE & BF16 & 1{,}048{,}576 (1M) & 52 \\
    \bottomrule
  \end{tabular}%
  }
  \end{small}
  \end{center}
\end{table*}

\paragraph{Training data.} OpenPerfectBlend \citep{labonneopenperfectblend} supplies an instruction
mixture spanning general conversation, mathematics, and code. For GPT-OSS-120B, assistant turns
are regenerated through SGLang with reasoning effort set to \texttt{medium} and temperature
$0.7$. The regenerated conversations are chat-templated and converted to Arrow. A teacher-forced
pass through the completed sequences caches input ids, the training loss mask, and target hidden
states at \texttt{target\_layer\_ids}, using sequence length $24{,}576$. The cache is reused across
teacher training, trajectory collection, and student distillation. 

\paragraph{Drafter architecture.} The GPT-OSS-120B drafter is an injection-conditioned Qwen3-style
transformer with $\block=16$, mask token id $200{,}000$, and
$\texttt{target\_layer\_ids}=[1,9,17,25,33]$. The backbone has $8$ decoder layers, hidden size
$2880$, intermediate size $7680$, and grouped-query attention with $64$ query heads and $8$
key/value heads. At each layer, a linear projection of the concatenated target features
(input dimension $5\times2880$), followed by RMSNorm, supplies additional attention keys and
values. The query stream carries the current block's token embeddings, including mask embeddings
at unrevealed positions. Target embeddings and
the LM head are frozen and shared; only draft-specific parameters are optimized. The configuration
estimate is approximately $0.8$B drafter parameters; the trainable parameter count is
$785$M. Drafter hidden dimensions and selected target layers for Gemma-4-31B,
Gemma-4-26B-A4B, and Nemotron-3.5-30B-A3B are respectively $5376$ ($[1, 12, 23, 35, 46, 57]$), $2816$ ($[1, 6, 11, 17, 22, 27]$), and $2688$ ($[1, 11, 20, 30, 39, 49]$).

% \paragraph{Drafter architecture.} The GPT-OSS-120B drafter is an injection-conditioned Qwen3-style
% transformer with $\block=16$, mask token id $200{,}000$, and
% $\texttt{target\_layer\_ids}=[1,9,17,25,33]$. The backbone has $8$ decoder layers, hidden size
% $2880$, intermediate size $7680$, and grouped-query attention with $64$ query heads and $8$
% key/value heads. At each layer, a linear projection of the concatenated target features
% (input dimension $5\times2880$), followed by RMSNorm, supplies additional attention keys and
% values. The query stream carries the current block's token embeddings, including mask embeddings
% at unrevealed positions. Target embeddings and
% the LM head are frozen and shared; only draft-specific parameters are optimized. The configuration
% estimate is approximately $0.8$B drafter parameters; the trainable parameter count is
% \textcolor{red}{XX}. Drafter dimensions and selected target layers for Gemma-4-31B,
% Gemma-4-26B-A4B, and Nemotron-3.5-Lightning-30B-A3B are \textcolor{red}{XX}.

\begin{table*}[t]
\centering
\caption{Architectural configuration of every draft model used in this paper, extracted directly from each model's \texttt{config.json}. \textit{Target Hidden Layers} lists the indices of the target model's layers whose hidden states are injected into the drafter; \textit{Attention} denotes sliding-window (SWA, with window size as subscript) vs.\ full attention over the drafted block. ``(Ours)'' rows are checkpoints we trained for DFlash and DSpark these are our own training runs of the published method on a backbone not released by its original authors. DStill is our proposed distillation method.}
\setlength{\tabcolsep}{4pt}
\resizebox{\textwidth}{!}{
\begin{tabular}{ll c cc l l}
\toprule
Model & Method & Params (M) & Hidden & Layers & Target Hidden Layers & Attention \\
\midrule
\multirow{2}{*}{GPT-OSS-120B}

& DFlash                           & 785 & 2880 & 8 & 1, 9, 17, 25, 33            & Full \\
& DStill (Ours)                          & 785 & 2880 & 8 & 1, 9, 17, 25, 33            & Full \\
\midrule
\multirow{2}{*}{Gemma-4-31B}

& DFlash                             & 1536 & 5376 & 5 & 1, 12, 23, 35, 46, 57     & SWA$_{2048}$+1 Full \\

& DStill (Ours)                            & 1536 & 5376 & 5 & 1, 12, 23, 35, 46, 57     & SWA$_{2048}$+1 Full \\
\midrule
\multirow{2}{*}{Gemma-4-26B-A4B}

& DFlash                                & 430  & 2816 & 5 & 1, 6, 11, 17, 22, 27 & SWA$_{2048}$ \\
% & DSpark                                & 564  & 2816 & 5 & 1, 6, 11, 17, 22, 27 & SWA$_{2048}$ \\
& DStill (Ours)                                & 430  & 2816 & 5 & 1, 6, 11, 17, 22, 27 & SWA$_{2048}$ \\
\midrule
\multirow{2}{*}{Nemotron-3.5-30B-A3B}
& DFlash                                     & 481 & 2688 & 6 & 1, 11, 20, 30, 39, 49 & SWA$_{2048}$ \\

% & DSpark                                    & 615 & 2688 & 6 & 1, 11, 20, 30, 39, 49 & SWA$_{2048}$ \\
& DStill (Ours)                                    & 481 & 2688 & 6 & 1, 11, 20, 30, 39, 49 & SWA$_{2048}$ \\
\bottomrule
\end{tabular}
}

\label{tab:draft_configs_summary}
\end{table*}

\paragraph{Training configuration.}\label{sec:gpu_compute}
Our training implementation builds on SpecForge
\citep{li2026specforge}. For GPT-OSS-120B, Stage I trains
the teacher for 10 epochs on 16 A100 nodes, and Stage II
distills the student for 3 epochs on 4 H100 nodes.
Both stages use a learning-rate warmup ratio
of $0.03$ and gradient-norm clipping at $1.0$.
Trajectory collection uses $K=8$ passes with greedy token
selection and retains the top-$16$ probabilities.
Trajectory-quality weighting is disabled except in its
dedicated ablation.

\paragraph{Benchmarks and generation settings.}
We evaluate 128 prompts each from GSM8K, MATH-500, and
MBPP; 164 each from HumanEval and SWE-bench; and 80 from
MT-Bench. Generation uses a maximum of 4096 new tokens,
with temperatures and reasoning settings specified in
the corresponding tables. GPT-OSS-120B uses reasoning
effort \texttt{medium}. Autoregressive and speculative
runs use the same requests and generation settings.
Benchmark averages are unweighted arithmetic means
across the six datasets, computed separately for speedup
and acceptance length. In the concurrency tables,
Avg~$\tau$ averages acceptance length over
$C\in\{1,8,16,32\}$.

\paragraph{Inference implementation.}
We use 8 H100 GPUs with vLLM version 0.30.0
\citep{kwon2023efficient}, tensor-parallel size 8,
and GPU memory utilization 0.8.
One-pass DFlash and \DStill evaluations use
\texttt{num\_denoising\_steps}$=1$ and
\texttt{num\_speculative\_tokens}$=15$.
The refinement comparison additionally evaluates
drafters with eight denoising passes.
DFlash and DSpark use the corresponding
\texttt{dflash} and \texttt{dspark} speculative
configurations.

\paragraph{Measurement.}
Each inference configuration is evaluated once after
four warmup runs, each allowing up to 128 generated
tokens. Generation time per output token is computed
by dividing the measured generation time by the actual
number of generated tokens.

\paragraph{Teacher-gain recovery and ablation controls.}
We compute recovery from acceptance lengths averaged
across the six benchmarks:
\[
\mathrm{Recovery}\,(\%)
=100\,
\frac{\tau_{\mathrm{student}}-\tau_{\mathrm{teacher},1}}
     {\tau_{\mathrm{teacher},8}-\tau_{\mathrm{teacher},1}}.
\]
Recovery is reported when the denominator is positive.
The state-sampling ablation uses Gemma-4-26B-A4B,
with both variants trained for 16 epochs on one-eighth
of the training corpus and evaluated on all six
benchmarks at concurrency 16. All other training
settings are held fixed.
Root-only sampling uses $p_{\mathrm{root}}=1$.
The cosine schedule increases this probability from
$0.5$ to $0.85$, with warmup and cooldown fractions
of $0.10$ and $0.30$, respectively. Its
training-averaged root probability is approximately
$0.71$.

% Requires the following packages in your preamble:
% \usepackage{booktabs}
% \usepackage{multirow}
% \usepackage{graphicx}

% \input{figure1}

\section{Reproducibility}
\label{sec:reproducibility}

Appendices~\ref{sec:prelim-details}
and~\ref{sec:training-details} detail the drafting,
verification, and training procedures.
Section~\ref{sec:exp-setup} and
Appendix~\ref{sec:exp-details} describe the datasets,
model configurations, hyperparameters, hardware,
and evaluation settings.

\section{Application to DSpark}
\label{appendix:dspark}
We apply trajectory consistency distillation to DSpark
\citep{cheng2026dspark}, jointly training its backbone
and auxiliary heads. We collect $K=8$ teacher refinement
trajectories and distill a one-pass student, retaining
the Markov and confidence losses in both training stages.
Table~\ref{tab:dsparkplusdstill} reports the per-benchmark
acceptance lengths and generation speedups.

\begin{table}[H]
  \centering
  \small
  \setlength{\tabcolsep}{3pt}
  \caption{Mean acceptance length $\tau$ and generation
  speedup $S$ over autoregressive decoding for DSpark
  and DSpark+\DStill. Average is the unweighted mean
  across the six benchmarks.}
  \label{tab:dsparkplusdstill}
  \begin{tabular*}{\columnwidth}
    {@{\extracolsep{\fill}}lrrrr@{}}
    \toprule
    & \multicolumn{2}{c}{\textbf{Acceptance length} $\boldsymbol{\tau}$}
    & \multicolumn{2}{c}{\textbf{Speedup} $\boldsymbol{S}$} \\
    \cmidrule(lr){2-3}
    \cmidrule(lr){4-5}
    \textbf{Dataset}
    & \textbf{DSpark} & \textbf{+\DStill}
    & \textbf{DSpark} & \textbf{+\DStill} \\
    \midrule
    GSM8K     & 6.74 & \textbf{6.89} & 4.23 & \textbf{4.29} \\
    HumanEval & 5.28 & \textbf{5.68} & 3.42 & \textbf{3.66} \\
    MATH-500  & 4.84 & \textbf{4.97} & 3.14 & \textbf{3.22} \\
    MBPP      & 5.01 & \textbf{5.11} & 3.25 & \textbf{3.29} \\
    MT-Bench  & 3.31 & \textbf{3.51} & 2.19 & \textbf{2.33} \\
    SWE-Bench & 3.03 & \textbf{3.28} & 2.02 & \textbf{2.19} \\
    \midrule
    \textbf{Average}
      & 4.70 & \textbf{4.91} & 3.04 & \textbf{3.16} \\
    \bottomrule
  \end{tabular*}
\end{table}

\end{document}